\documentclass{article}

\usepackage[final, nonatbib]{neurips_2024}

\usepackage[utf8]{inputenc} 
\usepackage[T1]{fontenc}    
\usepackage{hyperref}       
\usepackage{url}            
\usepackage{booktabs}       
\usepackage{amsfonts}       
\usepackage{nicefrac}       
\usepackage{microtype}      
\usepackage{xcolor}         

\usepackage{arydshln}
\usepackage{amsmath}
\usepackage{enumitem}
\usepackage{amssymb}
\usepackage{xspace}
\usepackage{mathtools}
\usepackage{amsthm}
\usepackage{caption}
\usepackage{subcaption}
\usepackage{algorithm}
\usepackage{algorithmic}
\usepackage{listings}
\usepackage{multirow}
\usepackage{multicol}
\usepackage[normalem]{ulem}
\usepackage{mdframed}
\usepackage{subfiles}
\usepackage{soul}
\usepackage{wrapfig}
\usepackage{titletoc}
\usepackage{tikz}
\usetikzlibrary{shadows}

\definecolor{myblue}{HTML}{1f77b4}
\definecolor{myorange}{HTML}{ff7f0e}
\definecolor{mygreen}{HTML}{ff7f0e}
\definecolor{colorGray}{HTML}{808080}
\definecolor{myred}{HTML}{d62728}
\definecolor{mypurple}{HTML}{9467bd}

\newcommand{\approach}{LoPA\xspace}
\usepackage{enumitem}

\theoremstyle{plain}

\title{Prompt Tuning Strikes Back: Customizing Foundation Models with Low-Rank Prompt Adaptation}

\author{%
  Abhinav Jain \\
  Department of Computer Science\\
  Rice University\\
  \texttt{aj70@rice.edu} \\
  \And
  Swarat Chaudhuri \\
  Department of Computer Science\\
  UT Austin \\
  \texttt{swarat@cs.utexas.edu} \\
  \And
  Thomas Reps \\
  Department of Computer Science\\
  University of Wisconsin-Madison \\
  \texttt{reps@cs.wisc.edu} \\
  \And
  Chris Jermaine \\
  Department of Computer Science\\
  Rice University \\
  \texttt{cmj4@rice.edu} \\
}

\begin{document}

\maketitle

\begin{abstract}
  Parameter-Efficient Fine-Tuning (PEFT) has become the standard for customising Foundation Models (FMs) to user-specific downstream tasks. However, typical PEFT methods require storing multiple task-specific adapters, creating scalability issues as these adapters must be housed and run at the FM server. Traditional prompt tuning offers a potential solution by customising them through task-specific input prefixes, but it under-performs compared to other PEFT methods like LoRA. To address this gap, we propose \textbf{Lo}w-Rank \textbf{P}rompt \textbf{A}daptation (LoPA), a prompt-tuning-based approach that performs on par with state-of-the-art PEFT methods and full fine-tuning while being more parameter-efficient and not requiring a server-based adapter. LoPA generates soft prompts by balancing between sharing task-specific information across instances and customization for each instance. It uses a low-rank decomposition of the soft-prompt component encoded for each instance to achieve parameter efficiency. We provide a comprehensive evaluation on multiple natural language understanding and code generation and understanding tasks across a wide range of foundation models with varying sizes. 
  
\end{abstract}

\section{Introduction}

Language models exhibit remarkable few-shot learning capabilities, demonstrating strong performance across tasks, including those unseen during training \cite{brown2020language, team2024gemma, llama3}. Nevertheless, fine-tuning remains crucial for optimised performance on a given downstream task. However, it becomes increasingly challenging with larger models because updating all parameters is impractical. Parameter Efficient Fine-Tuning (PEFT) presents a promising solution, adjusting a limited subset of parameters while leaving the rest unchanged. This approach allows the personalisation of pre-trained Foundation Models (FMs) to multiple users simultaneously.

\begin{wrapfigure}{r}{0.5\linewidth}
    \centering
    \includegraphics[width=\linewidth]{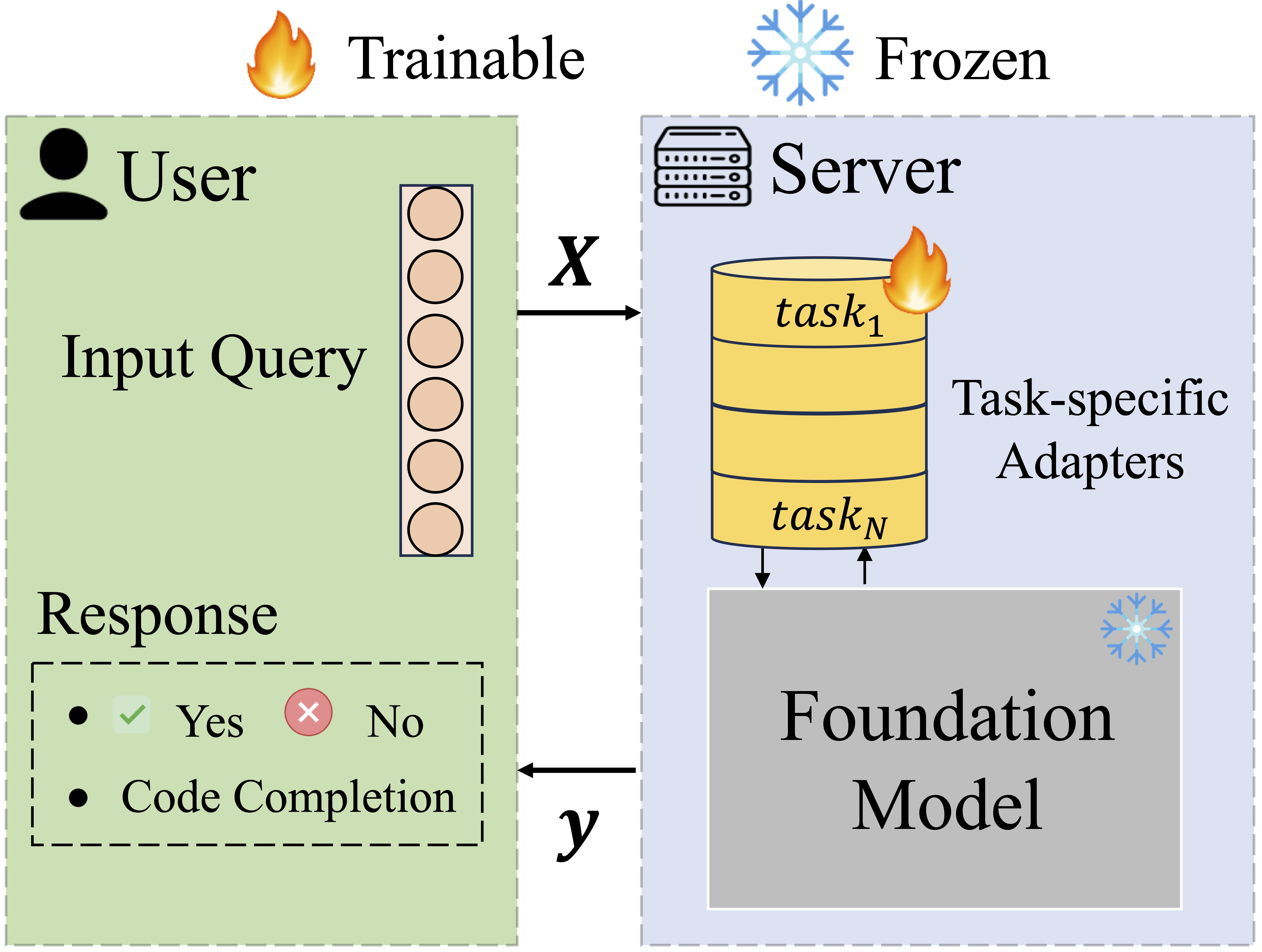}
    \caption{A schematic illustrating how typical PEFT methods like LoRA achieve personalization of a foundation model for multiple tasks, such as Yes/No text classification or code completion, during inference.}
    \label{fig:peft}
\end{wrapfigure}

In recent years, numerous PEFT approaches have been proposed \cite{houlsby2019parameter, hu2021lora, li2021prefix,  liu2022few, liu2024dora, shi2023dept,wen2024batched}, each varying in how and what they modify within FMs \cite{he2021towards, han2024parameter}. These variations can be categorized by the position in the FMs where modifications are applied, the functions used for modification, and the methods of integrating the modifications. While effective, these methods require maintaining multiple adapter-like modules for each user-specific task on the FM server and the need to select and assemble a subset of these modules every time a batch of user requests is processed for inference \cite{wen2023batched} (see Fig. \ref{fig:peft}).


Prompt tuning  \cite{lester2021power} is a simple approach that has certain advantages. First, it is parameter-efficient, requiring only a small set of vectors (soft prompts) that are prepended at the input layer and learned for a specific task. Second, prompt-based personalization requires no task-specific processing on the server. A task-specific prefix is added before processing, allowing the FM server to perform the same processing regardless of the task. However, despite its advantages, prompt tuning has been shown to underperform compared to other methods for PEFT \cite{he2021towards}. This gap in performance raises concerns about the viability of prompt tuning as a solution. Recently, efforts have been made to enhance the performance of prompt tuning by strategically inserting soft prompts into different layers of the transformer \cite{liu2021p, wu2022idpg, liu2022late, zhu2023spt}. However, this improvement increases the number of parameters and again necessitates server-side modifications for serving multiple tasks. An interesting recent work explored a simple method to make soft prompts input-dependent by using a lightweight prompt generator for each sample \cite{wu2022idpg}. This approach achieved notable improvements, raising the question: \textit{Can we further improve the performance of prompt tuning while staying parameter-efficient?}

To this end, we propose Low-rank Prompt Adaptation (\approach), a new instance-aware prompt tuning-based approach \footnote{The code for \approach can be found \href{https://github.com/jabhinav/Prompt-Tuning-Strikes-Back-with-LOPA}{here}}. \approach constructs the soft prompt from two components: a task-specific element that shares task information across samples and an instance-specific element that incorporates information from each individual instance. Our extensive analysis reveals that relying solely on either component, as done in previous works \cite{lester2021power, wu2022idpg}, is insufficient for outperforming other PEFT baselines. \approach achieves its high performance by taking a more balanced approach.


\approach combines the task-specific and instance-specific components using a gating function, which activates task-specific information conditioned on each instance. Additionally, it employs a low-rank decomposition of the instance-level component to enhance parameter efficiency (see Fig. \ref{fig:approach}). Once trained, users can provide the learned soft prompts as a prefix with their input to the FM, incurring no additional computational cost at the server.

We conducted extensive experiments on various models. These included six benchmark NLU tasks from the GLUE dataset and three Code Understanding and Generation tasks. Our results show that \approach outperforms existing prompt-tuning methods. It often matches the performance of full fine-tuning and LoRA. In 11 out of 24 test cases, we found \approach outperformed LoRA.

To summarize, the contributions of this work are as follows:
\begin{itemize}[leftmargin=0.2in]
    \item We propose \approach, a parameter-efficient and high-performing prompt-tuning strategy.
    \item We verify its effectiveness by evaluating it against full fine-tuning and state-of-the-art PEFT methods in nine tasks using seven different transformer backbones.
\end{itemize}
\section{Related Work}
\textbf{Adapter-based.} Several recent approaches have emerged to support parameter-efficient fine-tuning. These methods typically involve incorporating trainable adapter layers or modules into the transformer network \cite{houlsby2019parameter, he2021towards, liu2022few}. Training such layers has been demonstrated to be computationally more economical than full fine-tuning while maintaining performance. Notably, LoRA \cite{hu2021lora} has gained prominence for its low-rank approximation of model updates, effectively capturing task-specific knowledge. Beyond LoRA, Compacter \cite{karimi2021compacter} introduces adapters parameterised by low-rank hyper-complex multiplication  (PHM) layers \cite{zhang2021beyond} to achieve a more optimal balance between task performance and the number of trainable parameters. DoRA \cite{liu2024dora} is another recent approach that decomposes weights into their magnitude and directional components and employs LoRA for directional updates to minimise the number of trainable parameters efficiently.

\textbf{Soft Prompting.} Another line of work advocates for strategically inserting soft prompts into hidden states of the transformer instead of using trainable adapter modules. For example, prefix-tuning \cite{li2021prefix} and P-tuning-v2 \cite{liu2021p} prepend trainable prefix vectors to keys and values matrices at all transformer layers. Prompt tuning, P-tuning, and DePT \cite{lester2021power, liu2023gpt, shi2023dept} are special cases that operate by prepending prompt embeddings to the input at the first layer, with \cite{shi2023dept} being a hybrid-approach that further uses LoRA to learn updates for the input's embedding matrix. While these approaches are instance-agnostic and optimise a task-specific prompt, there are also methods to optimise an instance-aware soft prompt. For instance, IDPG \cite{wu2022idpg} generates a soft prompt for each sample, prepended either at the initial word-embedding layer (version-S) or all layers (version-M). LPT \cite{liu2022late} inserts a prompt into some intermediate layer (i) to eliminate gradient calculation below it for faster training and (ii) to retain more task-related information (which could be lost if it had to be propagated through lower layers). SPT \cite{zhu2023spt} aims to be more intelligent by learning a gating function to determine whether the soft prompt from the previous layer should be propagated or if a new one should be learned.

With the exception of prompt tuning and S-IDPG, PEFT approaches mostly operate by injecting prefixes and new trainable modules into deeper layers or doing low-rank re-parameterization of existing ones, necessitating the storage of PEFT parameters at the server to update the foundation model. In contrast, \approach provides the soft prompt as a prefix that is prepended to the input query, overcoming the need to store task-specific parameters on the server. Furthermore, we demonstrate that \approach achieves a more balanced trade-off between specificity and generalization compared to existing approaches for enhancing learned models, such as prompt tuning \cite{lester2021power}, which solely focuses on a general task-specific soft prompt, and IDPG \cite{wu2022idpg}, which emphasizes an instance-specific prompt.
\section{Proposed Methodology}
In this section, we formally define the proposed approach, followed by an explanation of how it affects model learning as compared to existing soft-prompting approaches.

\subsection{Preliminaries}

\textbf{Transformers.}
A transformer model consists of multiple stacked layers, where each layer has multiple heads ($=N_H$), each performing self-attention over the input \cite{vaswani2017attention}. Let us consider a single head, $H$ parameterised by the query, key, and value weights as $W^Q, W^K, W^V  \in \mathcal{R}^{d_H \times d}$, respectively, where $d$ is the model dimension. In multi-headed attention, $d_H$ is typically set to $\frac{d}{N_H}$. For a given sequence of $n$ input vectors $\textbf{X} = \{\textbf{x}^1, \ldots, \textbf{x}^n \}$ and a query vector $\textbf{x}^i$ with each $\textbf{x}\in \mathcal{R}^d$ , the output of the head at position $i$ is -
\begin{equation}
    \textbf{o}^i = \textrm{Attention}(W^Q \textbf{x}^i, W^K\textbf{X},W^V\textbf{X}) = \text{softmax}\big(W^Q \textbf{x}^i(W^K\textbf{X})^\top\big)W^V\textbf{X} \label{eq: xformer_op}
\end{equation}
 where, we have ignored the constant $\sqrt{d_H}$ for notational convenience. For the first layer of the transformer network, the input $\textbf{X}$ is the embedding matrix i.e. $\textbf{X} = \textbf{X}_e \in \mathcal{R}^{d\times n}$.

\textbf{Full Fine-Tuning (FFT).} With fine-tuning, the objective is to adapt the model to a new task with data $\mathcal{D}=\{(\textbf{X}, \textbf{y})\}$. Formally, adaptation is achieved by updating the weights $(W)$ of the model to maximise the log-likelihood of the response $\textbf{y}$, i.e., 
$\max_{W} \mathcal{L} = \mathbb{E}_{(\textbf{X}, \textbf{y}) \sim \mathcal{D}}[\log p_{W}(\textbf{y}|\textbf{X})]$.


\textbf{Prompt-Tuning (PT).} The objective of prompt-tuning is to achieve task-specific model adaptation by learning a "soft prompt" $\textbf{Z}$. Such a soft prompt is prepended to the input word embeddings and trained via back-propagation: $\max_{\textbf{Z}} \mathcal{L} = \mathbb{E}[\log p_{W}(\textbf{y}|\mathrm{concat}(\textbf{Z}, \textbf{X}_e))]$. In PT,  the soft prompt $\textbf{Z}\in \mathcal{R}^{d\times m}$ is parameterised by a set of $m$ learnable vectors $\{\textbf{z}^1, \ldots \textbf{z}^m\}$, where $m$ denotes its length \cite{lester2021power}. Thus, it can be interpreted as an embedding of virtual tokens that enable the model to perform a downstream task without updating its parameters \cite{qin2021learning, hambardzumyan2021warp}.

With prompt tuning, the output of an attention head in the first layer is modified as follows:
\begin{equation*}
    \textbf{o}_{PT}^i = \textrm{Attention}(W^Q \textbf{x}^i, W^K \textrm{concat}(\textbf{Z},\textbf{X}_e), W^V \textrm{concat}(\textbf{Z},\textbf{X}_e))
\end{equation*}
By using the formulation of \textrm{Attention} from Eq. \ref{eq: xformer_op}, this equation can be equivalently written as
\begin{align}
    \textbf{o}_{PT}^i &= \underbrace{\sum_k A_{ik}W^V\textbf{z}^k}_{bias} + \underbrace{(1-\sum_k A_{ik})}_{scale}\textbf{o}^i \label{eq: bias}\\
    \text{with}\quad A_{ik} &= \frac{\exp((W^K\textbf{z}^k)^\top W^Q \textbf{x}^i)}{\sum_k\exp((W^K\textbf{z}^k)^\top W^Q \textbf{x}^i)) + \sum_j\exp((W^K\textbf{x}^j)^\top W^Q \textbf{x}^i))} \label{eq: attention}
\end{align}
where, $A_{ik}$ is the attention transformer gives to the prefix vector $\textbf{z}_k$ for a given query vector $\textbf{x}^i$. In Eq. \ref{eq: bias}, we can observe that the soft prompt linearly interpolates the head's position-wise output; where the bias term can be considered in an offset subspace spanned by vectors $\{W^V\textbf{z}^k\}_{k=1}^m$ with dimension $m$ (or $\leq m$)  \cite{he2021towards}.

\subsection{Low-rank Prompt Adaptation (LoPA)}


\approach constructs the soft prompt as \(\textbf{Z} = \textbf{Z}_{S} \circ g(\textbf{Z}_{I})\), where \(\textbf{Z}_{S} \in \mathbb{R}^{d \times m}\) is the task-specific component and \(\textbf{Z}_{I} \in \mathbb{R}^{d \times m}\) is the instance-specific component. These are combined using the gating function \(g\), implemented using sigmoid, where \(\circ\) denotes the Hadamard product. Intuitively, by sharing \(\textbf{Z}_S\) across instances, it captures general information, adapting the model to user-defined tasks, while \(\textbf{Z}_I\) fine-tunes the soft prompt for specific instances, acting as activations.
\begin{wrapfigure}{r}{0.65\linewidth}
    \centering
      \includegraphics[width=\linewidth]{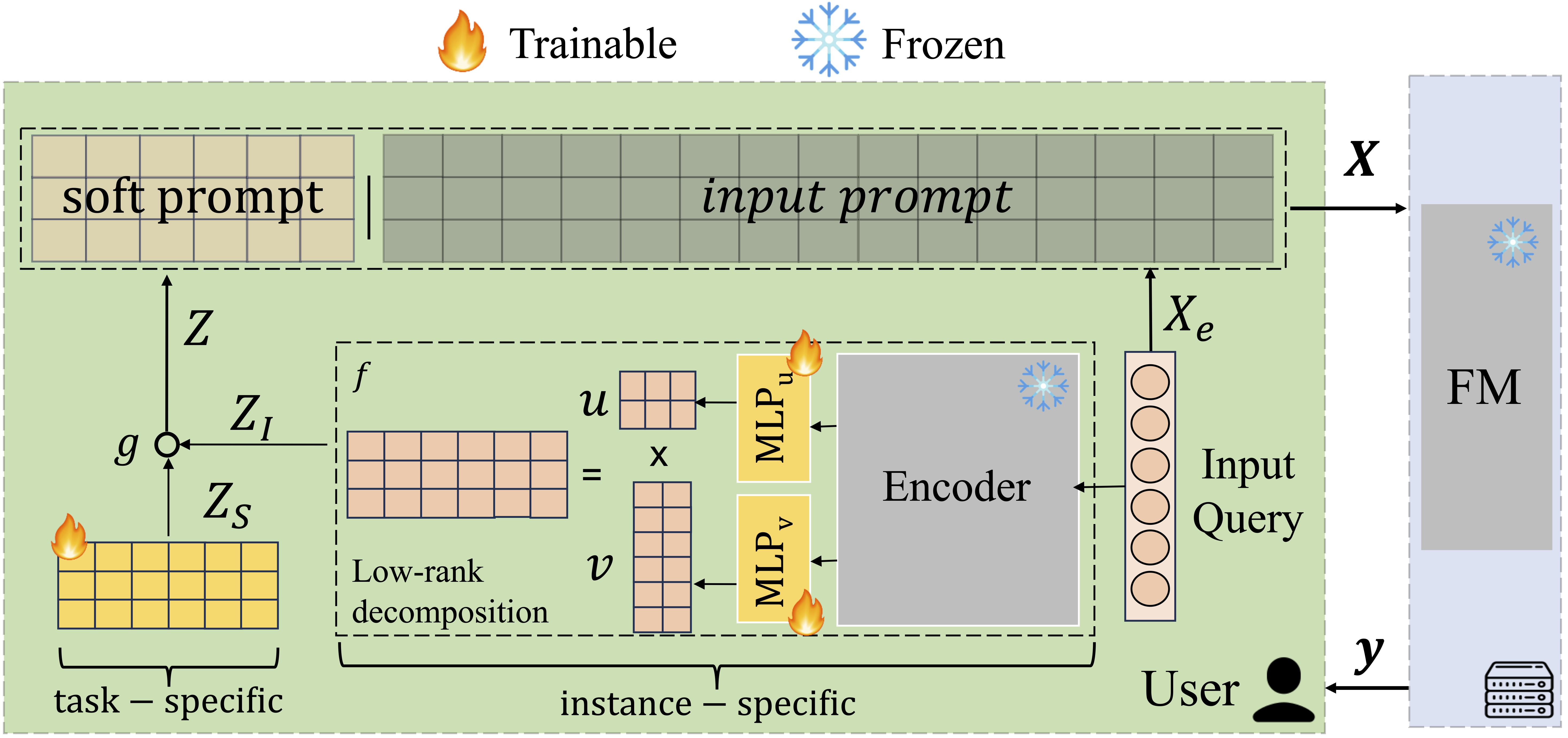}
    \caption{An illustration of \approach. No task-specific adapters need to be stored on the server. $|$ represents the concatenation of the soft prompt \(\textbf{Z}\) and the input prompt \(\textbf{X}_e\) i.e. $\textbf{X}=\textrm{concat}(\textbf{Z}|\textbf{X}_e)$}
    \label{fig:approach}
\end{wrapfigure}
Both \(\textbf{Z}_S\) and \(\textbf{Z}_I\) have their own dedicated parameters. Similar to prompt tuning, \(\textbf{Z}_S\) consists of \(m\) learnable vectors. Conversely, \(\textbf{Z}_I\) is obtained from input using the encoding function \(f : \mathbb{R}^{d \times n} \to \mathbb{R}^{d \times m}\). However, encoding a matrix of size \(d \times m\) can be expensive. For example, in \cite{wu2022idpg}, an MLP layer with hidden dimension \(h\) requires \(\mathcal{O}(hdm)\) parameters. To improve parameter efficiency, we assume a low-rank decomposition of \(\textbf{Z}_I\) as $\textbf{Z}_I=\textbf{u}\times \textbf{v}^\top$ and encode the two matrices with rank $r$, \(\textbf{u} \in \mathbb{R}^{d \times r}\) and \(\textbf{v} \in \mathbb{R}^{m \times r}\), using separate MLP layers.
This design makes \(f\) computationally cheaper, with \(\mathcal{O}(hdm(\frac{r}{d}+\frac{r}{m}))\) parameters, reduced by a factor of \((\frac{r}{d}+\frac{r}{m}) < 1\) when $r$ is chosen to be \( \ll \min(d, m)\). The overall composition of \(\textbf{Z}\) can be represented as:
\begin{equation}
    \textbf{Z} = \underline{\textbf{Z}_S} \circ g(\underbrace{\underline{\texttt{MLP}_\textbf{u}}(\textbf{X}') \times \underline{\texttt{MLP}_\textbf{v}}(\textbf{X}')^\top}_{\textbf{Z}_I=f(\textbf{X}')})
    \label{eq:lopa}
\end{equation}
where \_ marks the three trainable components. The $\texttt{MLP}$ head consists of a down- and up-projection layer with a non-linear activation in between. It sits atop a smaller language model, referred to as the Encoder model, that gives the input representation $\textbf{X}'$. A visual illustration of the framework can be found in Fig. \ref{fig:approach}.


Next, we provide an intuitive explanation why \approach could be better than traditional prompt-tuning \cite{lester2021power} and existing instance-specific approaches \cite{wu2022idpg}.


\textbf{Offset subspace induced by \approach.} From Eq.2, we can observe that the bias vector in the offset subspace is a linear combination of $\{W^V\textbf{z}^k\} \quad \forall k \in \{1, \ldots, m\}$, with \(A_{ik}\) (Eq. 3) representing the scalars \cite{he2021towards, petrov2023prompting}. In \approach, the input influences \(\textbf{Z}\), causing the vectors $\{W^V\textbf{z}^k\}$ to vary accordingly. Consequently, distinct offset sub-spaces emerge for different inputs. In contrast, traditional prompt tuning maintains fixed vectors while allowing only the scalars to vary with input. As a result, instance-sensitive approaches can exert greater influence on the attention patterns in deeper layers of the transformer, thereby offering enhanced flexibility and responsiveness to varying inputs.


\textbf{Coupled learning of ${\textbf{Z}_S}$ and ${\textbf{Z}_I}$.} Let's examine the partial derivatives of the objective \(\mathcal{L}\) with respect to \(\textbf{Z}_S\) and \(\textbf{Z}_I\). Using the chain-rule on the formulation in Eq. \ref{eq:lopa} with $g(.)=\textrm{sigmoid}(.)$,
\begin{align*}
    \nabla_{\textbf{Z}_S}\mathcal{L}&=\nabla_{\textbf{Z}} \mathcal{L} \circ \sigma(\textbf{Z}_I)\\
    \nabla_{\textbf{Z}_I}\mathcal{L}&=(\nabla_{\textbf{Z}} \mathcal{L} \circ \textbf{Z}_S) \cdot \sigma(\textbf{Z}_I)(1 - \sigma(\textbf{Z}_I))
\end{align*}

These expressions suggest a coupled learning process where changes in \(\textbf{Z}_I\) (through the sigmoid function) directly impact the updates to \(\textbf{Z}_S\), and the updates to \(\textbf{Z}_I\) are scaled by both \(\textbf{Z}_S\) and the derivative of the sigmoid function. In contrast, consider IDPG with \(\textbf{Z} = \texttt{MLP}(\textbf{X}')\) \cite{wu2022idpg}. Here, the bias term of the MLP layer can be viewed as a task-specific element, such that the composition is \(\textbf{Z} = \textbf{Z}_S + \textbf{Z}_I\). This approach results in updates to \(\textbf{Z}_S\) and \(\textbf{Z}_I\) being independent of each other. Such a linear-sum operation may fail to capture the complex relationships between the two components that \approach can capture due to \approach's non-linear factorization.

\section{Experiments, Results, and Discussion}

\begin{table}
  \centering
  \renewcommand{\arraystretch}{1.2} 
  \captionsetup{skip=5pt} 
  \begin{tabular}{ccccccccc}
    \toprule
    Tuning & Tunable & SST-2 & MNLI & MRPC & QNLI & QQP & RTE & Avg \\
     & Parameters & (acc) & (acc) & (acc, F1) & (acc) & (acc, F1) & (acc) &  \\
    \midrule
    FFT & 355M & 95.99 & \textbf{90.40} & 90.81 & 94.60 & \textbf{90.39} & \textbf{85.92} & \textbf{91.35} \\
    p-tuning-v2 & 0.49M & 89.91 & 88.13 & 70.18 & 85.21 & 84.63 & 53.43 & 78.58 \\
    prefix-tuning & 25.75M & 93.80 & 89.51 & 90.86 & \textbf{94.98} & 86.87 & 82.67 & 89.78 \\
    LoRA & 2.36M & \textbf{96.22} & 90.30 & 90.77 & 94.69 & 89.91 & 85.66 & 91.26 \\
    \cdashline{1-9}
    None & 0 & 59.97 & 39.60 & 73.52 & 50.16 & 42.34 & 53.43 & 53.17 \\
    PT & 10.2K & 84.40 & 54.67 & 72.38 & 58.74 & 48.20 & 53.07 & 61.91\\
    DePT & 10.2K & 89.68 & 71.51 & 72.97 & 57.09 & 45.81 & 53.79 & 65.14 \\
    S-IDPG & 2.89M & 95.30 & 84.50 & 78.60 & 90.48 & 84.88 & 77.26 & 85.17 \\
    Ours & 1.60M & \underline{95.99} & \underline{89.22} & \underline{\textbf{91.09}} & \underline{93.74} & \underline{89.72} & \underline{83.39} & \underline{90.53} \\
    \bottomrule
  \end{tabular}
  \caption{Performance on GLUE tasks. We report the average of accuracy and F1 for both MRPC and QQP. For all the other tasks, we report accuracy. Approaches below the dotted line do not require any modification to the model on the server side. \textbf{Bold} denotes the best-performing tuning method for the given model. \underline{Underline} marks the best result among all prompt tuning methods.}
  \label{tab:glue}
\end{table}

\begin{table}
  \centering
  \renewcommand{\arraystretch}{1.2} 
  \captionsetup{skip=5pt} 
  \begin{tabular}{cccccc}
    \toprule
   \multirow{2}{*}{Model} & \multirow{2}{*}{Tuning} & \multirow{2}{*}{\#Params} & \multicolumn{2}{c}{Code Understanding} & Code Generation\\
   \cline{4-5}
    &  &  & CruxEval-I & CruxEval-O & MBPP \\
   \midrule
     
     \multirow{6}{*}{CodeGen-350M}
     & FFT & 350M & 33.0 & \textbf{19.5} & 17.49\\
     & LoRA & 1.3M & 31.0 & 18.2 & \textbf{21.56}\\
        \cdashline{2-6}
     & None & 0 & 20.8 & 15.0 & 15.85\\
     & PT & 10.2K & 32.8 & 15.8 & 15.20\\
     & S-IDPG & 8.5M & 16.9 & 13.2 & \underline{17.04}\\
     & Ours & 4.4M & \underline{\textbf{34.5}} & \underline{18.5} & \underline{17.04}\\
    
    \hline
    \rule{0pt}{3ex}
  
    \multirow{6}{*}{DeepseekCoder-1.3B} 
     & FFT & 1.3B & \textbf{45.0} & 34.8  & \textbf{44.76}\\
     & LoRA & 4.7M & 35.5 & \textbf{36.0} & 44.14\\
       \cdashline{2-6}
     & None & 0 & 26.8 & 29.8 & 34.08\\
     & PT & 20.5K & 41.2 & 31.2 & 34.49\\
     & S-IDPG & 16.3M & 26.0 & 28.5 & 42.50\\
     & Ours & 4.2M & \underline{43.0} & \underline{34.5} & \underline{44.66}\\
     
     \hline
     \rule{0pt}{3ex}
    
    \multirow{6}{*}{Phi2-2.7B} 
     & FFT & 2.7B & 40.2 & 37.0  & \textbf{55.03}\\
     & LoRA & 7.9M & 41.5 & \textbf{42.5} & 51.54\\
        \cdashline{2-6}
     & None & 0 & 33.5 & 33.0 & 45.17\\
     & PT & 25.6K & 35.0 & 34.0 & 49.69\\
     & S-IDPG & 20.3M & 35.0 & 33.0 & \underline{53.29}\\
     & Ours & 4.76M & \underline{\textbf{43.0}} & \underline{37.2} & 52.15\\
    
    \hline
    \rule{0pt}{3ex}
    
    \multirow{6}{*}{Phi3-3.8B}
     & FFT & 3.8B & 39.2 & 39.5 & \textbf{54.00}\\
     & LoRA & 6.3M & 38.0 & 41.2 & 42.92\\
      \cdashline{2-6}
     & None & 0 & 33.8 & 39.5 & 8.82\\
     & PT & 30.7K & 33.5 & 31.5 & 34.08\\
     & S-IDPG & 24.2M & 31.0 & 39.5 & 42.29\\
     & Ours & 5.3M & \underline{\textbf{42.2}} & \underline{\textbf{42.5}} & \underline{44.35}\\
    
    \hline
    \rule{0pt}{3ex}
    
    \multirow{6}{*}{DeepseekCoder-7B} 
     & FFT & 7B & \textit{OOM} & \textit{OOM} & \textit{OOM}\\
     & LoRA & 11.8M & 47.5 & \textbf{49.8} & 53.38\\
            \cdashline{2-6}     
     & None & 0 & 39.3 & 44.0 & 50.51\\
     & PT & 41.0K & 45.8 & 44.8 & 37.47\\
     & S-IDPG & 32.1M & 40.5 & 41.5 & \underline{\textbf{53.59}}\\
     & Ours & 6.35M & \underline{\textbf{50.0}} & \underline{48.0} & 52.46\\
     
     \hline
     \rule{0pt}{3ex}
     
     \multirow{6}{*}{Llama3-8B}
     & FFT & 8B & \textit{OOM} & \textit{OOM} & \textit{OOM}\\
     & LoRA & 9.4M & \textbf{45.5} & \textbf{40.5} & 44.55\\
        \cdashline{2-6}
     & None & 0 & 27.0 & 31.5 & \textbf{45.37}\\
     & PT & 41.0K & 37.5 & 32.0 & 26.07\\
     & S-IDPG & 32.1M & 29.2 & 35.2 & 33.05\\
     & Ours & 6.4M & \underline{41.2} & \underline{39.8} & \underline{43.94}\\
     \bottomrule
  \end{tabular}
  
  \caption{Performance comparison on CruxEval and MBPP tasks. We report average $\textit{pass}@1$ scores. Approaches below the dotted line are prompt-tuning methods, which do not require any modification to the model on the server side. \textbf{Bold} denotes the best-performing tuning method for the given model. \underline{Underline} marks the best result among all prompt-tuning methods. \textit{OOM} indicates that the corresponding tuning approach exceeded the available GPU memory and ran out of memory.}

  \label{tab:code}
\end{table}

\subsection{Experimental Setup}

\textbf{Tasks.} We evaluate \approach on (i) six Natural Language Understanding (NLU) tasks from the GLUE benchmark \cite{wang2018glue}---namely, SST-2 \cite{socher2013recursive}, MNLI \cite{williams2017broad}, MRPC \cite{dolan2005automatically}, QNLI \cite{rajpurkar2016squad}, QQP, and RTE \cite{giampiccolo2007third}; (ii) a code-generation task that requires the model to complete method bodies from MBPP benchmark \cite{austin2021program}, and (iii) two code-understanding tasks---namely, CruxEval-I (input prediction) and CruxEval-O (output prediction) from CruxEval benchmark \cite{gu2024cruxeval}. These tasks assess the model's ability to reason about the execution of simple Python programs by asking it to predict the input given the output of a function and vice-versa.



\textbf{Baselines.} We evaluate \approach against the following baselines that represent different customisation approaches. (1) FFT, (2) LoRA \cite{hu2021lora}, (3) p-tuning-v2 \cite{liu2021p} and (4) prefix-tuning (h=512) \cite{li2021prefix} are selected as representatives of methods that customise models on the server side, with FFT representing full fine-tuning and the remainder showcasing parameter-efficient techniques. (5) Standard prompt-tuning \cite{lester2021power}, (6) S-IDPG \cite{wu2022idpg} and (7) DePT \cite{shi2023dept} are chosen because they customise models from the user side, focusing on soft prompt insertion with DePT also learning updates to the input embedding matrix. For IDPG, we use DNN layers in MLP-head to encode the soft prompt for the input embedding layer. We chose DNN over PHM layers \cite{zhang2021beyond} because we found them to exhibit better performance across our task set. Refer to Appendix \ref{subsec:PHMcomparison} for the comparison. Additionally, the proposed \approach also uses DNN layers, facilitating a fair comparison with S-IDPG-DNN \cite{wu2022idpg}. Lastly, we include results for the model without any fine-tuning to report its (8) zero-shot performance.

\textbf{Backbone Architectures.} For NLU tasks, we test all the tuning methods on 355M \texttt{RoBERTa}  \cite{liu2019roberta} similar to the setup opted by prior work \cite{liu2021p, wu2022idpg, zhu2023spt}. For Code Generation and Understanding, we test a subset of the baselines (FFT, LoRA, PT, S-IDPG) on a range of FM backbones; starting from smaller FMs: 350M \texttt{CodeGen-mono} \cite{nijkamp2022codegen} and 1.3B \texttt{Deepseek-Coder} \cite{guo2024deepseek}; mid-sized FMs: 2.7B \texttt{phi-2} \cite{javaheripi2023phi}, 3.8B \texttt{phi-3} \cite{phi3}; and larger FMs: 7B \texttt{Deepseek-Coder} \cite{guo2024deepseek} and 8B \texttt{Llama 3} \cite{llama3}.

\textbf{Implementation Details.} For the GLUE tasks, we use the train-test splits pre-defined in the benchmark, while for the MBPP and CruxEval tasks, we employ a 50-50 split. Across all tasks and backbone models, the soft prompting baselines are implemented with $m=10$ virtual tokens representing the soft prompt. For NLU tasks, both IDPG and \approach use \(\texttt{RoBERTa} + \texttt{MLP(h=256)}\) as the encoder, whereas in code tasks, \(\texttt{125M CodeSage} + \texttt{MLP(h=1024)}\) is used. Additionally, for \approach, the best-performing rank $(r)$ of the low-rank decomposition is chosen from \(\{1, 2, 4\}\), and the corresponding number of tunable parameters is reported.

\textbf{Training Configuration.} For NLU tasks, training with FFT and LoRA was done for 10 epochs, while with prompt-tuning-based approaches it was done for 20 epochs. In MBPP, all foundation model (FM) backbones were trained for 10 epochs across all tuning methods. In CruxEval Tasks across all PEFT methods, FM backbones under 7B were trained for 20 epochs, while larger FMs ($\geq$7B) were trained for 10 epochs. Lastly, training with FFT on CruxEval tasks was done for 5 epochs. The learning rates for \approach are set to \(1 \times 10^{-5}\) in NLU and \(1 \times 10^{-3}\) in Coding tasks. The baseline tuning methods use the following learning rates across all the tasks: FFT using \(1 \times 10^{-5}\), LoRA and the remainder of soft-prompting approaches using \(1 \times 10^{-4}\).  All experiments are conducted on 40GB 2xA100 GPUs. 

\textbf{Evaluation.} We report binary or multi-class classification accuracies and F1 scores for NLU tasks as provided in the GLUE benchmark. For coding tasks, we report the $\textit{pass}@1$ scores computed using the best-performing temperatures: $0.6$ for MBPP and $0.2$ for CruxEval.

\begin{figure}[t]
    \centering
    \begin{subfigure}[b]{0.49\linewidth}
        \centering
        \includegraphics[width=\linewidth]{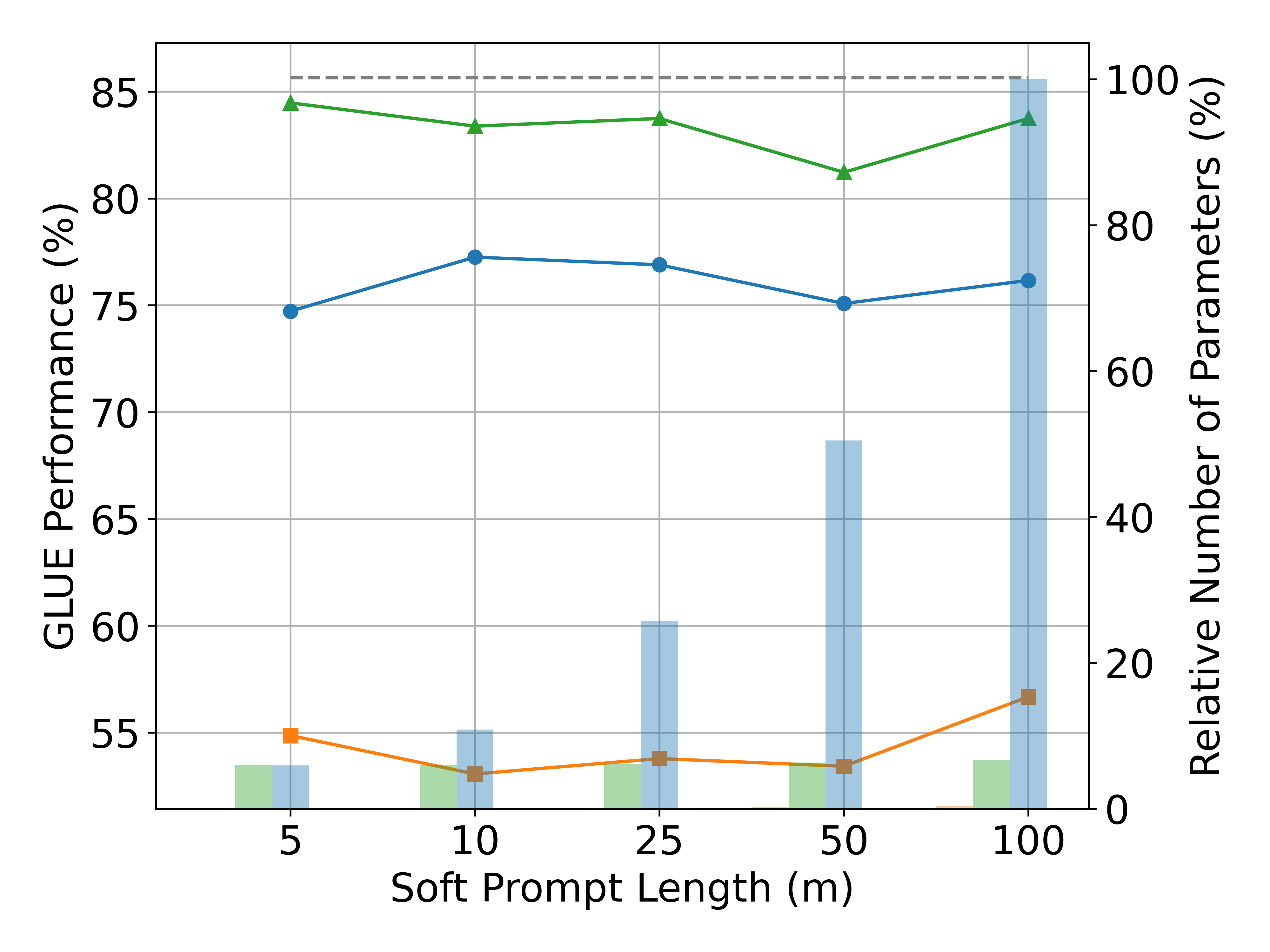}
        \caption{RTE}
        \label{subfig:rte_baseline}
    \end{subfigure}
    \hfill
    \begin{subfigure}[b]{0.49\linewidth}
        \centering
        \includegraphics[width=\linewidth]{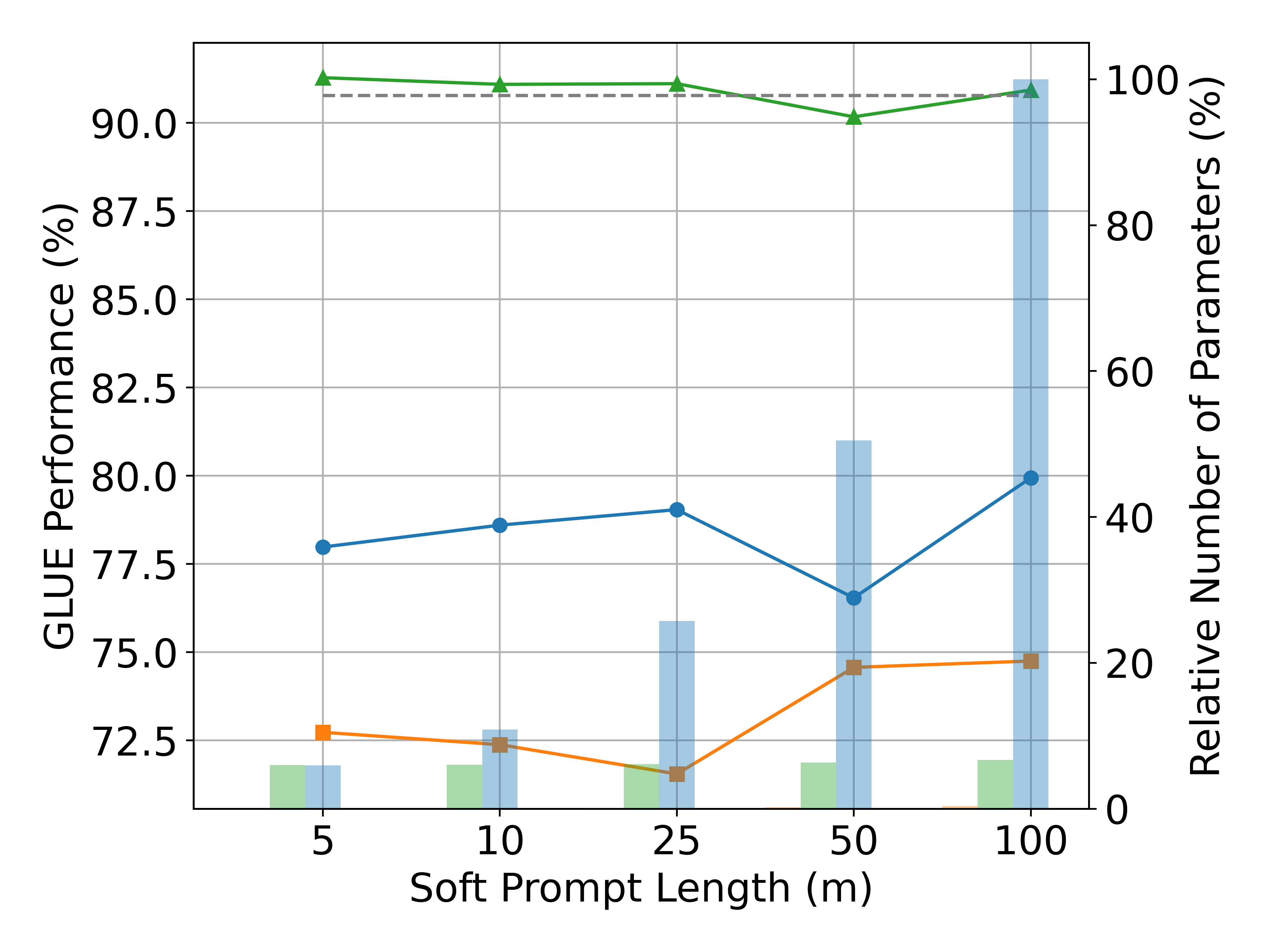}
        \caption{MRPC}
        \label{subfig:mrpc_baseline}
    \end{subfigure}
    \begin{subfigure}[b]{0.49\linewidth}
        \centering
        \includegraphics[width=\linewidth]{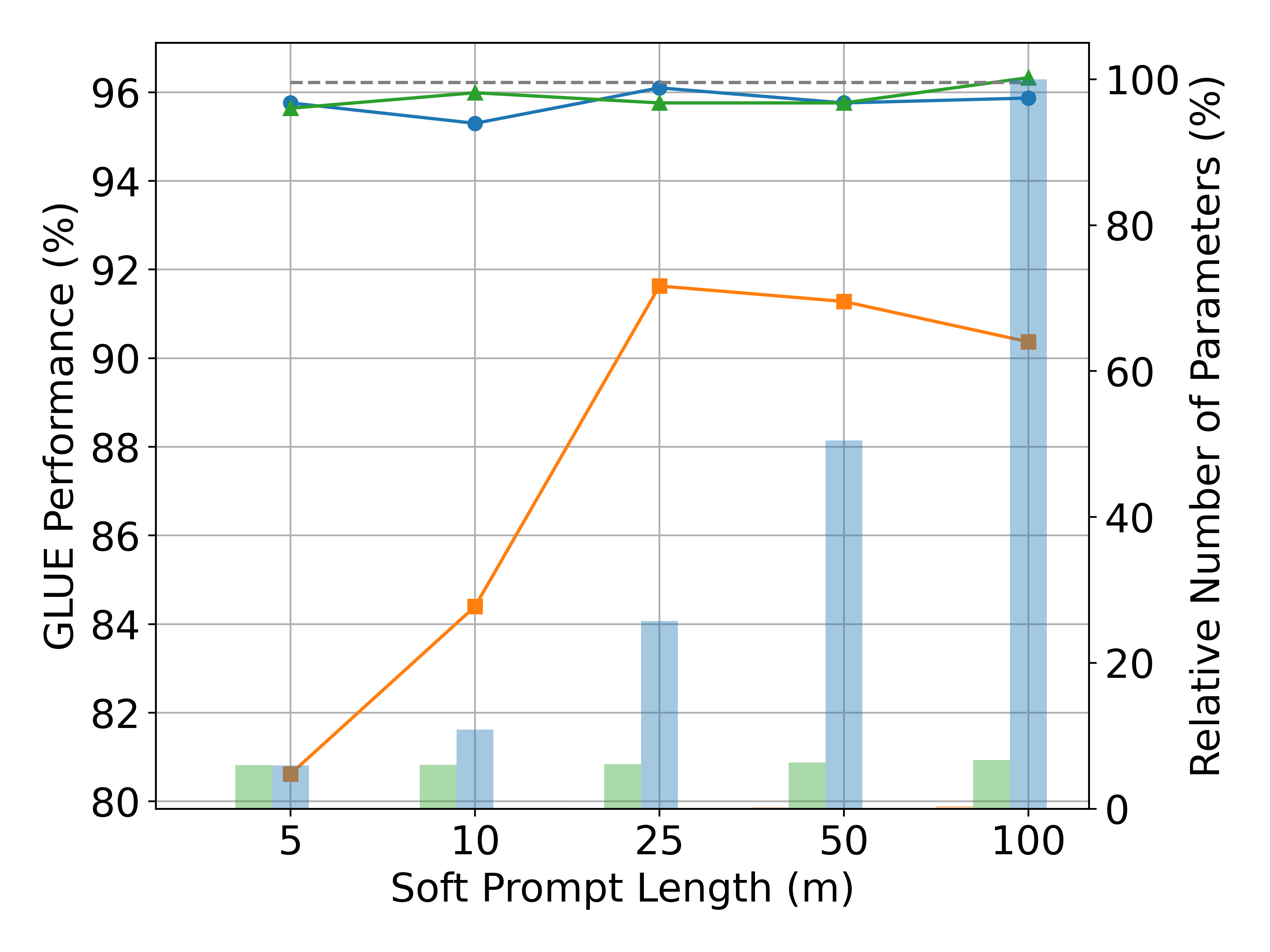}
        \caption{SST-2}
        \label{subfig:sst2_baseline}
    \end{subfigure}
    \hfill
    \begin{subfigure}[b]{0.49\linewidth}
        \centering
        \includegraphics[width=\linewidth]{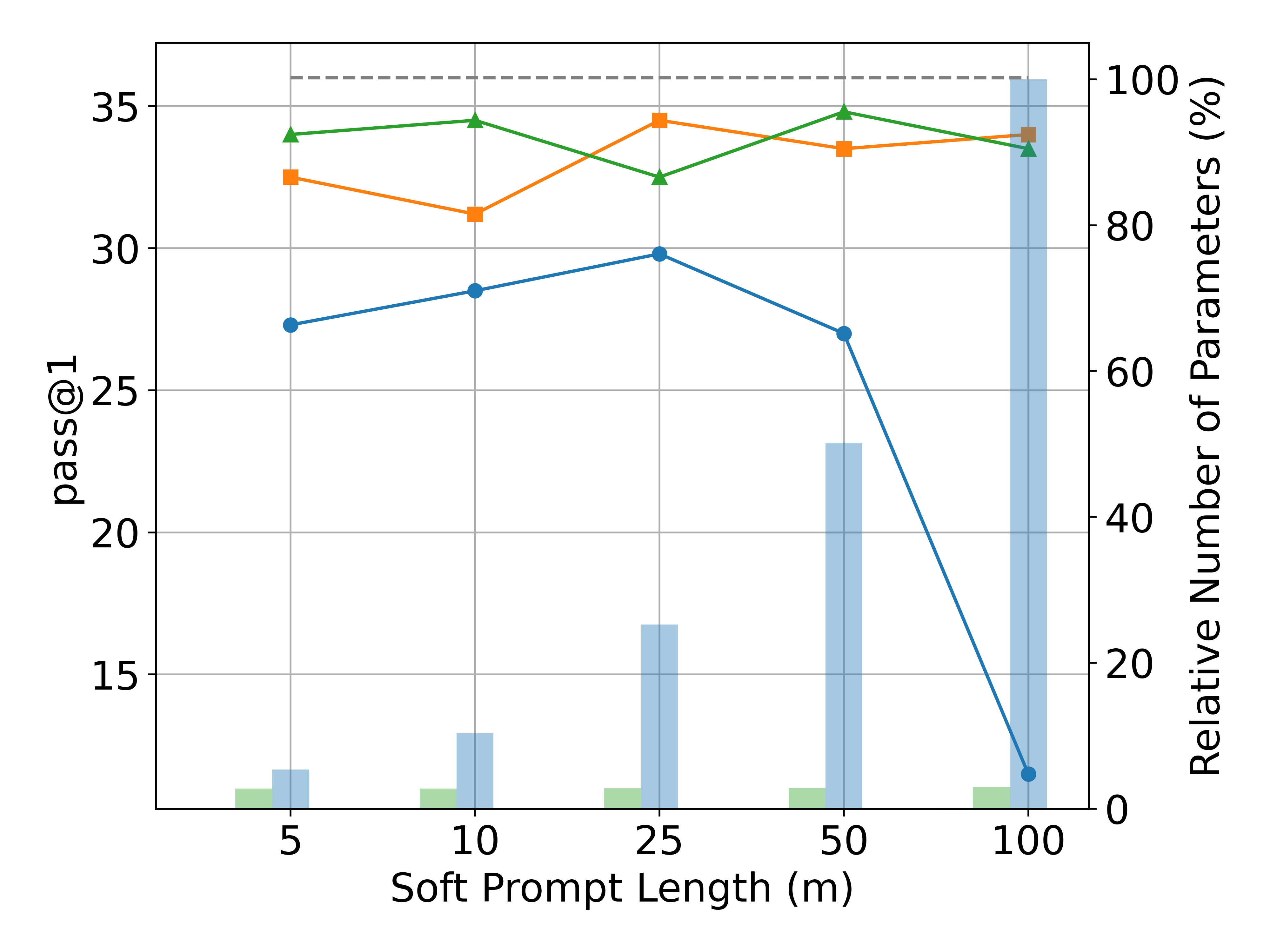}
        \caption{CruxEval-O}
        \label{subfig:cruxeval-o_baseline}
    \end{subfigure}
    \vskip\baselineskip
    \centering
    \begin{tabular}{ll}
        \includegraphics[width=6cm]{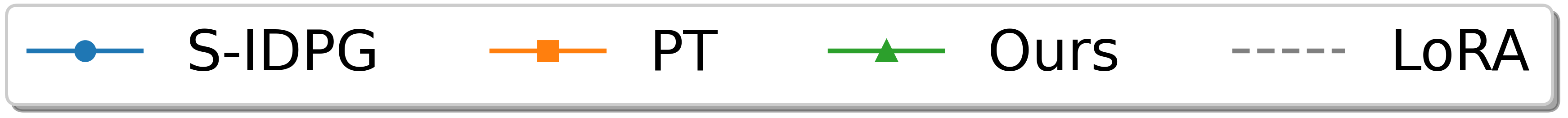} &  \includegraphics[width=6cm]{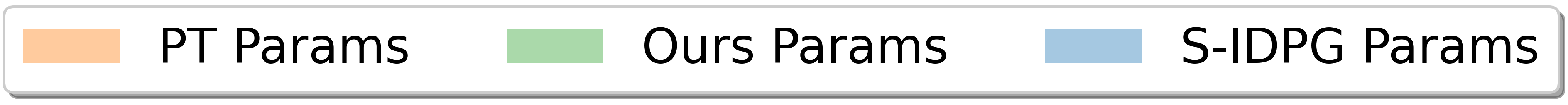}
    \end{tabular}
    \caption{Performance comparison of baselines as a function of $m$ on (a)-(c) GLUE benchmark and (d) CruxEval-O (with DeepseekCoder-1.3B as FM). Tunable parameters shown relative to the method with the most. Higher performance and fewer parameters indicate better results.}
    \label{fig:ablation_prompt_length}
\end{figure}

\subsection{Baseline Comparison}

\textbf{Performance on Natural Language Understanding.} In Table \ref{tab:glue}, we can observe that the \approach consistently outperforms the traditional prompt-tuning method and DePT by an average margin of $28.62$ points and $25.39$ points respectively. This result demonstrates that conditioning the soft prompt on instances enables it to influence the model’s output more significantly. Furthermore, \approach shows an average improvement of $5.36$ points over IDPG, highlighting that the proposed factorisation captures complex relationships between task-specific (\(\textbf{Z}_S\)) and instance-specific (\(\textbf{Z}_I\)) components. This improvement is particularly notable in limited-data settings, with a $12.5$-point increase in MRPC and a $6.13$-point increase in RTE.

Additionally, \approach achieves performance close to FFT and LoRA, within $1$ point, while using $760k$ fewer parameters than LoRA owing to \approach's low-rank decomposition of the soft prompt. This performance profile suggests that the \approach is a parameter-efficient and high-performing alternative for NLU tasks. It outperforms existing prompt-tuning approaches and performs on par with FFT and LoRA, making it a compelling choice for efficient and effective model tuning.


\begin{table}[t]
  \centering
  \renewcommand{\arraystretch}{1.2} 
  \captionsetup{skip=5pt} 
  \begin{tabular}{lccccccc}
    \toprule
    $\textbf{Z}=$ & SST-2 & MNLI & MRPC & QNLI & QQP & RTE & Avg \\
     & (acc) & (acc) & (acc, F1) & (acc) & (acc, F1) & (acc) &\\
    \midrule
    $\textrm{concat}(\textbf{Z}_S, \textbf{Z}_I)$ & 94.50 & 78.45 & 76.00 & 91.43 & 76.11 & 84.12 & 83.44 \\
    $\max(\textbf{Z}_S, \textbf{Z}_I)$ & \textbf{95.99} & \textbf{89.37} & 88.62 & \textbf{93.74} & 78.44 & \textbf{85.92} & 88.68 \\
    $\textbf{Z}_{S}\circ g(\textbf{Z}_{I})$ & \textbf{95.99} & 89.22 & \textbf{91.09} & \textbf{93.74} & \textbf{89.72} & 83.39 & \textbf{90.53} \\
    \bottomrule
  \end{tabular}
  \caption{Performance of \approach on GLUE tasks with respect to function encoding $\textbf{Z}$. $\textrm{concat}(.)$ represents the concatenation of $\textbf{Z}_{S}$ and $\textbf{Z}_{I}$. $\textrm{max}(.)$ represents the element-wise max operation. We report the average of accuracy and F1 for both MRPC and QQP. For all the other tasks, we report accuracy. \textbf{Bold} denotes the best-performing encoding function for \approach.}
  \label{tab:ablation_composition_glue}
\end{table}

\textbf{Performance on Code Understanding.} In Table \ref{tab:code}, \approach consistently improves the $\textit{pass}@1$ score of the baseline with no tuning across all FM backbones. It outperforms prompt-tuning on CruxEval tasks, with modest improvements of approximately $2$ to $4$ points on smaller FMs like CodeGen and DeepSeekCoder-1.3B and larger improvements ranging from $8$ to $11$ points on larger FMs like LLama-3 and Phi-3 in CruxEval-O. Furthermore, IDPG performs worse than PT for all models except Phi-3 in CruxEval-O. These results suggest that merely encoding an instance-sensitive soft prompt does not guarantee improvements and can even degrade performance (e.g., IDPG on CodeGen and DeepSeekCoder-1.3B in CruxEval tasks). The poor generalization of the learned soft prompt, possibly due to over-parameterization and resulting overfitting, might explain this behaviour.
In contrast, \approach, being more parameter-efficient, explicitly incorporates task and instance information in its design of the soft prompt, leading to better performance.


\approach also performs on par with LoRA, often within a range of $1$ to $4$ points of $\textit{pass}@1$, while roughly using two-thirds of the parameters. Notably, \approach outperforms LoRA across all models in CruxEval-I, except for LLama-3, with improvements approximately ranging from $2$ points in DeepSeekCoder-7B to $4$ points in Phi-3. This result might be attributed to the fact that many of the FMs considered here are good knowledge approximators, well-trained on diverse datasets, and demonstrate strong zero-shot generalization. Directly updating a subset of their weights can still lead to catastrophic forgetting, where the models lose previously acquired knowledge \cite{pfeiffer2020adapterfusion}. In such cases, soft prompting, as employed by \approach, can effectively elicit the necessary skills to solve new tasks without compromising existing knowledge \cite{petrov2023prompting}.

\textbf{Performance on Code Completion.} On the code completion task of MBPP, both IDPG and \approach improve the performance of the baseline model with nearly equal gains except for LLama-3. However, \approach achieves this with significantly fewer tunable parameters—approximately half the number used in CodeGen and only one-fifth of those used by IDPG in DeepseekCoder-7b. This demonstrates that \approach scales well with the size of the foundation model, maintaining both performance and parameter efficiency. This efficiency is attributed to the low-rank approximation of the instance-specific matrix employed by \approach. For LLama-3, all tuning approaches led to a drop in performance, possibly due to over-fitting, suggesting that LLama-3 might already be trained on MBPP.

\textbf{Overview of Results.} Averaged over all tasks and foundation models, \approach showed relative percentage improvements of \(28.52\%\) over PT and \(20.16\%\) over IDPG, while being outperformed by LoRA by only \(0.54\%\). Notably, \approach outperformed LoRA in 11 out of 24 cases. Thus, in the test cases we considered, there was no clear systematic advantage to LoRA in terms of accuracy. Given that \approach requires no task-specific processing at the server—the prompt can be computed anywhere, even at the client, before being sent to the server for processing—we believe \approach may be a useful alternative to LoRA for some tasks.


 \begin{figure}
     \centering
     \begin{tabular}{cc}
        \includegraphics[width=0.49\linewidth]{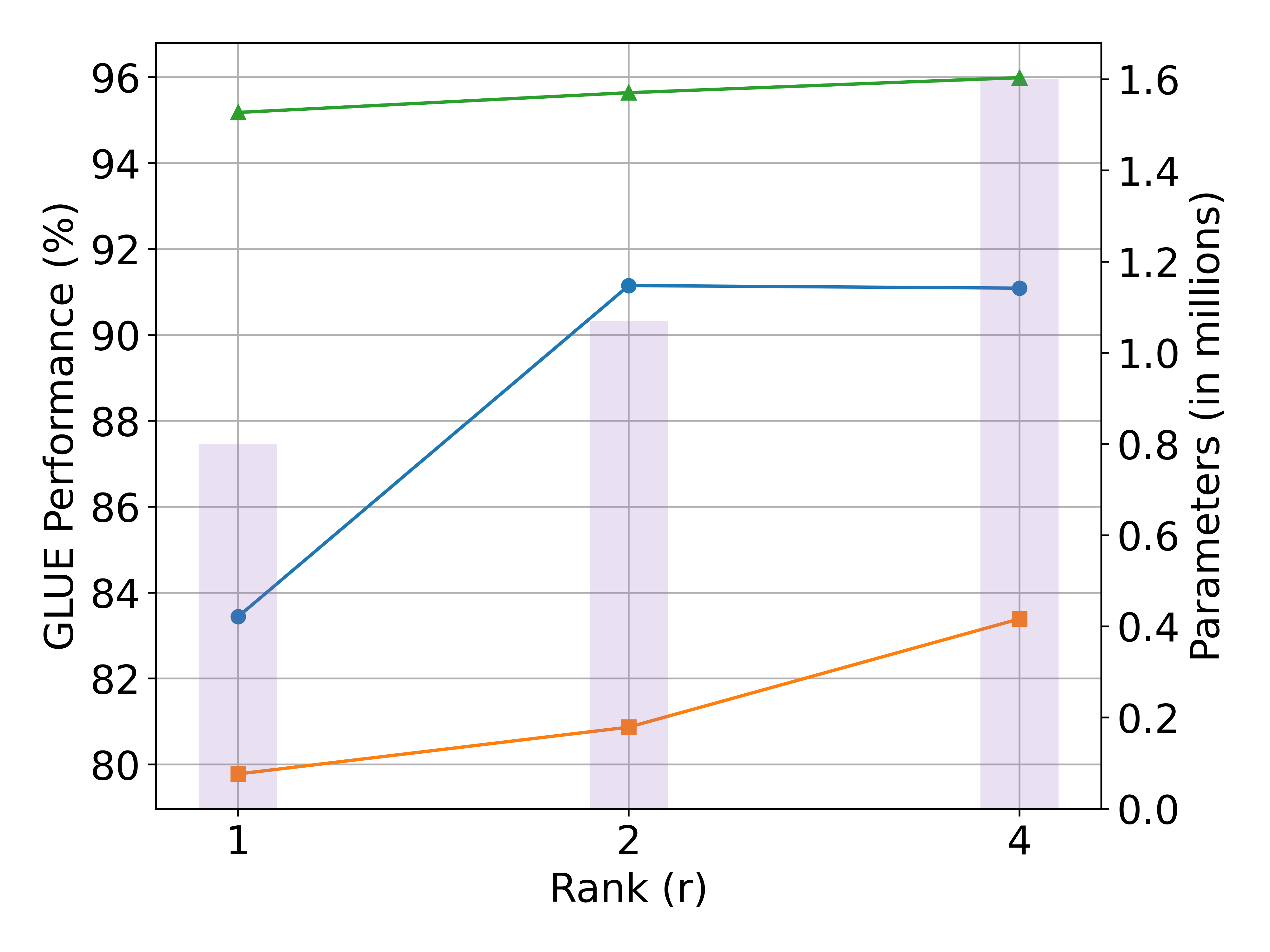}  &  \includegraphics[width=0.49\linewidth]{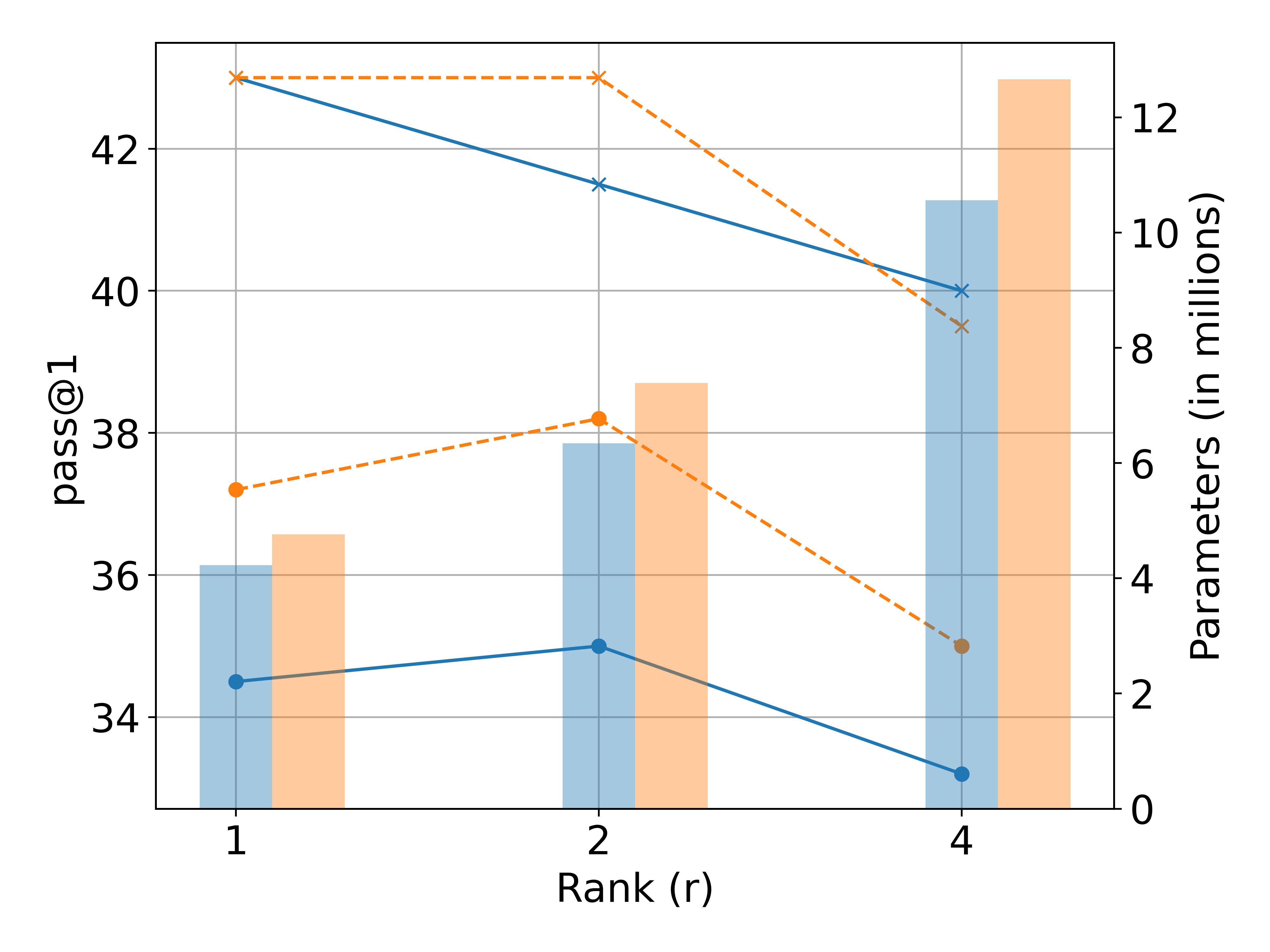}\\
        \includegraphics[width=0.4\linewidth]{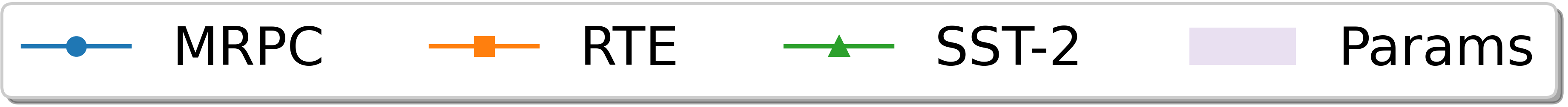}  &  \includegraphics[width=0.4\linewidth]{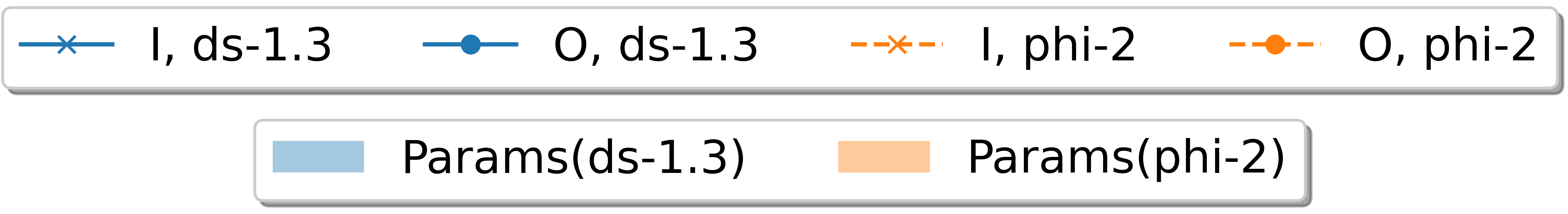}\\
        (a) NLU & (b) CruxEval
     \end{tabular}
     \caption{Performance of \approach as a function of rank shown for $m=10$. (a) GLUE Benchmarks and (b) CruxEval tasks $(I, O)$ where ds-1.3 denotes DeepseekCoder-1.3B and phi-2 denotes Phi2-2.7B models. Higher performance and fewer tunable parameters indicate better results.}
     \label{fig:ablation_rank}
 \end{figure}

\subsection{Ablation Study}
\textbf{Performance of \approach as a function of soft-prompt length.} In Figure \ref{fig:ablation_prompt_length}, we study how the length of the soft prompt impacts the performance of \approach compared to other prompting methods. Increasing the length of the soft prompt corresponds to adding more vectors to the set that represents the soft prompt, thus expanding the offset subspace (See Eq. \ref{eq: bias}). Whether the added vectors are mutually independent and provide additional useful information depends on the tuning approach to learning them and the offset subspace of the FM.

For instance, PT and IDPG initially see performance improvements with increased prompt length, but performance eventually plateaus or drops due to over-fitting (See PT on SST-2 Fig. \ref{subfig:sst2_baseline} and IDPG on CruxEval-O Fig. \ref{subfig:cruxeval-o_baseline}). In contrast, \approach does not exhibit significant performance fluctuations with varying prompt lengths (See Fig. \ref{subfig:rte_baseline}-\ref{subfig:cruxeval-o_baseline}). This stability is likely due to the shared component \(\textbf{Z}_S\) acting as a regularizer, preventing over-fitting of the instance-specific soft prompts.

Moreover, \approach with \(m=5\) outperforms PT and IDPG even when they use longer prompts (\(m > 5\)) (Refer Fig. \ref{subfig:rte_baseline}, \ref{subfig:mrpc_baseline}). This result suggests that the dimension of the offset subspace is much smaller, and \approach can more accurately represent it with its learned vectors.

\textbf{Performance of \approach as a function of rank.} In Figure \ref{fig:ablation_rank}, we examine how the rank of the proposed low-rank decomposition of the instance-specific component affects the performance. For NLU, we consider three tasks: MRPC, RTE, and SST-2, and observe the performance of \texttt{RoBERTa} as the rank increases from 1 to 4 . Performance consistently improves with increasing rank, showing gains of $1\%$ to $2 \%$ for SST-2 and up to $8\%$ for MRPC.
\begin{wrapfigure}{r}{0.5\linewidth}
    \centering
    \begin{tabular}{|c|c|c|}
        \hline
        Encoder & CruxEval-I & CruxEVal-O\\
        \hline
        CodeBert(125M) & 41.2 & 32.8 \\
        \hline
        CodeSage(130M) & \textbf{43.0} & \textbf{34.5} \\
        \hline
        CodeSage(365M) & 42.2 & \textbf{34.5} \\
        \hline
    \end{tabular}
    \caption{Ablation for Encoder in \approach with DeepseekCoder-1.3B  as the foundation model.}
    \label{fig:enc_ablation}
\end{wrapfigure}
In contrast, for CruxEval tasks, increasing the rank does not proportionally improve the performance. We attribute this behaviour to the size of the datasets used to approximate the low-rank matrices. NLU tasks provide thousands of samples, allowing for better approximation of higher-order matrices. However, CruxEval has only a few hundred samples, and increasing the rank introduces more parameters, possibly leading to over-fitting.

\textbf{Performance of \approach as a function of encoder network.}
In Figure \ref{fig:enc_ablation}, we study the impact on the performance by choosing different transformer backbones for the Encoder network in \approach. For this experiment, we use $125M$ \texttt{CodeBERT} \cite{feng2020codebert}, and $130M$ and $365M$ \texttt{CodeSAGE} \cite{zhang2024code} encoder models to generate input encodings, $\textbf{X}'$ for the CruxEval tasks. We observe that \texttt{CodeSAGE} models achieve a $2$-point improvement over CodeBERT in both tasks. This improvement can be attributed to \texttt{CodeSAGE}'s superior pre-training using Contrastive Learning, which allows for finer distinctions in code representations. Consequently, the soft-prompt vectors $\{\textbf{z}^k\}$ in \approach, as functions of the input \(\textbf{X}'\), capture instance-specific nuances more effectively and accordingly exert influence on the model's output. We also tried tuning the encoder model while learning soft prompts but did not find any significant improvements in the performance. 

\textbf{Performance of \approach with respect to function encoding $\textbf{Z}$.} In Table \ref{tab:ablation_composition_glue}, we evaluate the effect of different functions encoding $\textbf{Z}$ as a combination of $\textbf{Z}_I$ and $\textbf{Z}_S$ on RoBERTa’s performance in NLU tasks. The results show that simply concatenating $\textbf{Z}_S$ and $\textbf{Z}_I$ performs the worst while the non-linear functions, such as max(.) and the proposed gating mechanism, yield the best performance. We leave the exploration of other non-linear functions for future work.




\section{Conclusion}
In this paper, we introduced Low-Rank Prompt Adaptation (\approach), an instance-specific soft-prompting method that outperforms other methods in the prompt-tuning family, and performs on par with full fine-tuning and LoRA, while using fewer tunable parameters. \approach first uses a low-rank approximation of the instance-specific soft prompt and combines it with a task-specific soft prompt via a gating function. With a more informed way of designing soft prompts, this work aims to position prompt tuning as a powerful alternative to adapter-based methods for user-specific customization of foundation models.

\textbf{Limitations and Future Work.} The main limitation of \approach is that its effectiveness as an alternative to LoRA was demonstrated on a set of benchmark tasks, but this may not hold for obscure real-life user tasks where LoRA or even full fine-tuning might be necessary. Further investigation into its performance on real-world tasks is part of our future work. In this work, we assumed the learned soft prompt to be prepended as a prefix to the input. Future research could explore the effects of positioning it as a suffix or randomly within the input. Finally, the foundation model combined with \approach can be viewed as a Conditional Auto-Encoder, where soft prompt vectors exist in a latent subspace rather than an offset subspace. This viewpoint raises intriguing questions, such as whether the observed performance improvements result from inferring and compressing task-specific knowledge and providing it as additional information. Investigating this alternate perspective could lead to further performance enhancements by developing more sophisticated auto-encoding systems. 

\textbf{Broader Impact.} Our contribution to new knowledge is the development of a language-model customization method that delivers strong performance while being parameter-efficient. The significance of our work lies in its potential to reduce training and maintenance costs associated with hosting and customizing foundation models. Furthermore, our method enhances user privacy by enabling task-specific customization on the user end rather than the server end.



\section{Acknowledgments}
This research was supported by the NSF under grant numbers CCF1918651, 2008240, 2131294, and 2212557, NIH CTSA award No. UL1TR003167, and US DOT Tier-1 UTC CYBER-CARE grant \#69A3552348332.

\small{
\bibliographystyle{plain}
\bibliography{neurips_2024}
}


\newpage
\section{Appendix}
\subsection{Comparison with Parameterized Hypercomplex Multiplication (PHM) layers}\label{subsec:PHMcomparison}

\begin{table}[h]
  \centering
  \renewcommand{\arraystretch}{1.2} 
  \captionsetup{skip=5pt} 
  \begin{tabular}{lcccccccc}
    \toprule
    Tuning & Tunable & SST-2 & MNLI & MRPC & QNLI & QQP & RTE & Avg \\
    Method & Params & (acc) & (acc) & (acc, F1) & (acc) & (acc, F1) & (acc) &  \\
    \midrule
    \approach & 1.60M & \textbf{95.99} & \textbf{89.22} & \textbf{91.09} & \textbf{93.74} & \textbf{89.72} & \textbf{83.39} & \textbf{90.53} \\
    S-IDPG &  &  &  &  &  &  & &  \\
    \; + FC & 2.89M & 95.30 & 84.50 & 78.60 & 90.48 & 84.88 & 77.26 & 85.17 \\
    \; + PHM (n=8) & 0.37M & 95.07 & 83.46 & 76.12 & 85.45 & 77.35 & 67.14 & 80.77 \\
    \; + PHM (n=16) & 0.20M & 94.61 & 83.66 & 76.68 & 81.16 & 80.39 & 65.34 & 80.31\\
    \; + PHM (n=32) & 0.17M & 94.72 & 81.45 & 74.99 & 84.59 & 81.90 & 68.23 & 80.98\\
    \bottomrule
  \end{tabular}
  \caption{Performance on GLUE tasks. We report the average of accuracy and F1 for both MRPC and QQP. For all the other tasks, we report accuracy. \textbf{Bold} denotes the best-performing tuning method for the given model. FC represents the fully-connected layers and PHM represents the hypercomplex multiplication layers parameterised by $n$.}
  \label{tab:phm_ablation}
\end{table}

In this section, we consider PHM layers as an alternative to low-rank decomposition in LOPA to reduce parameter complexity. We carry out an ablation study on NLU tasks where we implement IDPG with PHM layers as $\textbf{Z}_I = \texttt{PHM}_{W}(\textbf{X}')$ with $W=\sum_{i=1}^nA_i\bigotimes B_i$, where $\bigotimes$ represents the Kroneckr product between learnable matrices $A_i, B_i$ and $n$ represents the hyper-parameter balancing the parameter complexity and extent of factorisation in the Kronecker product. We can observe in Table \ref{tab:phm_ablation} that while PHM layers reduce parameters, they also lead to a significant performance drop of approximately $10$ points on average compared to LOPA. This drop may be due to the structural bias in $W$ imposed by Kronecker factorisation of PHM layers, which could limit expressiveness \cite{zhang2021beyond} in comparison to Fully-Connected layers of DNN. We want to point out that for a further reduction in trainable parameters, \approach can also be used in conjunction with PHM layers.

\subsection{Convergence Analysis of Soft-Prompting Approaches}
 
We present plots in Figures \ref{fig:qqp}-\ref{fig:qnli} comparing the training loss and performance on NLU tasks (QQP, QNLI, MNLI) for Prompt Tuning (PT), IDPG, and LOPA. The results show that instance-dependent methods like IDPG and LOPA converge faster than traditional prompt tuning. Moreover, LOPA converges faster and achieves higher accuracy or F1 scores compared to IDPG.

\definecolor{lightblue}{rgb}{0.68, 0.85, 0.9}  
\definecolor{red}{rgb}{1.0, 0.0, 0.0}          
\definecolor{darkgreen}{rgb}{0.0, 0.5, 0.0}    

\begin{figure}[H]
  \centering
  \subfloat[Training Loss]{
    \includegraphics[width=0.48\textwidth]{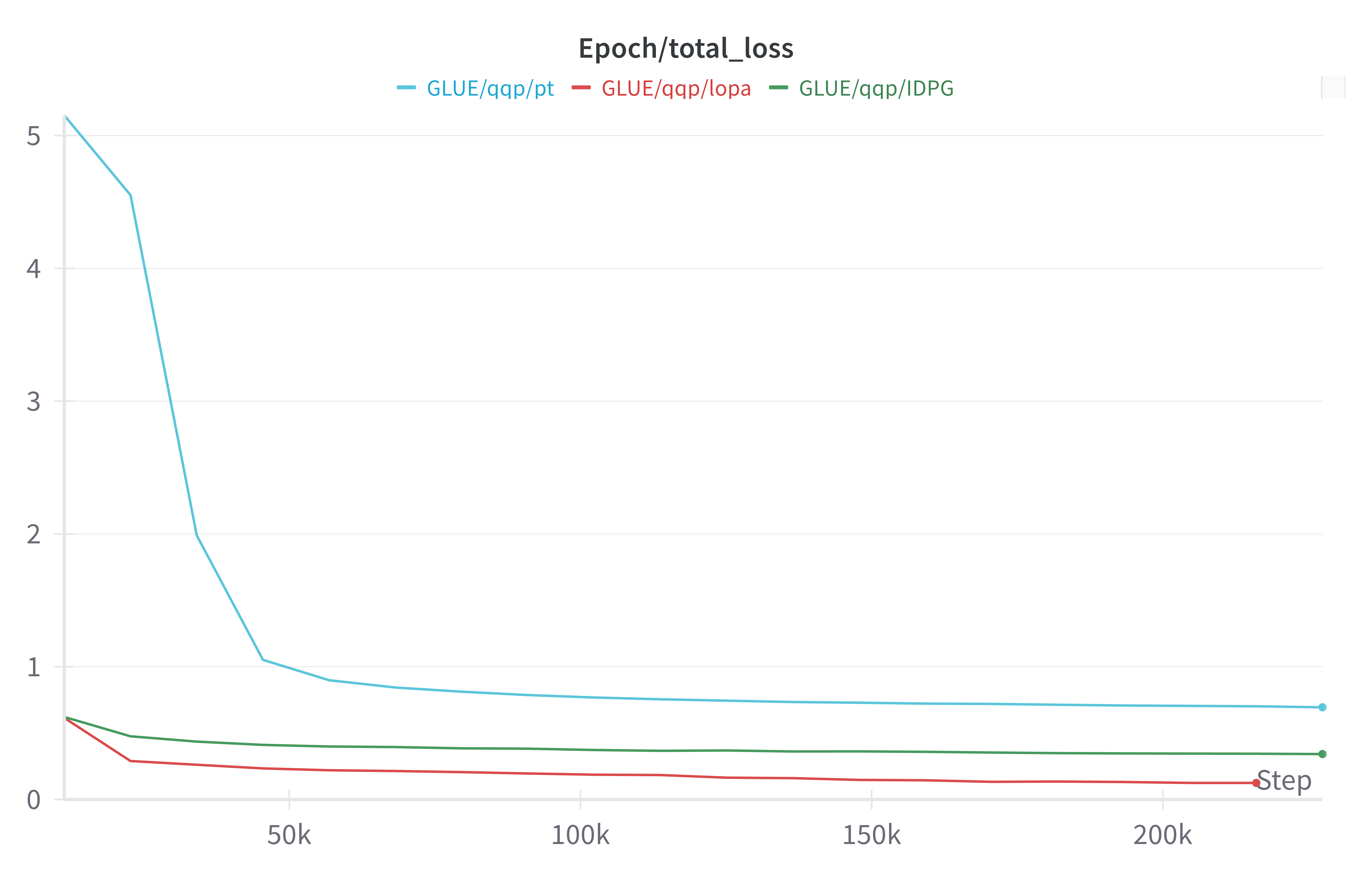}
  }
  \hfill
  \subfloat[Validation Accuracy]{
    \includegraphics[width=0.48\textwidth]{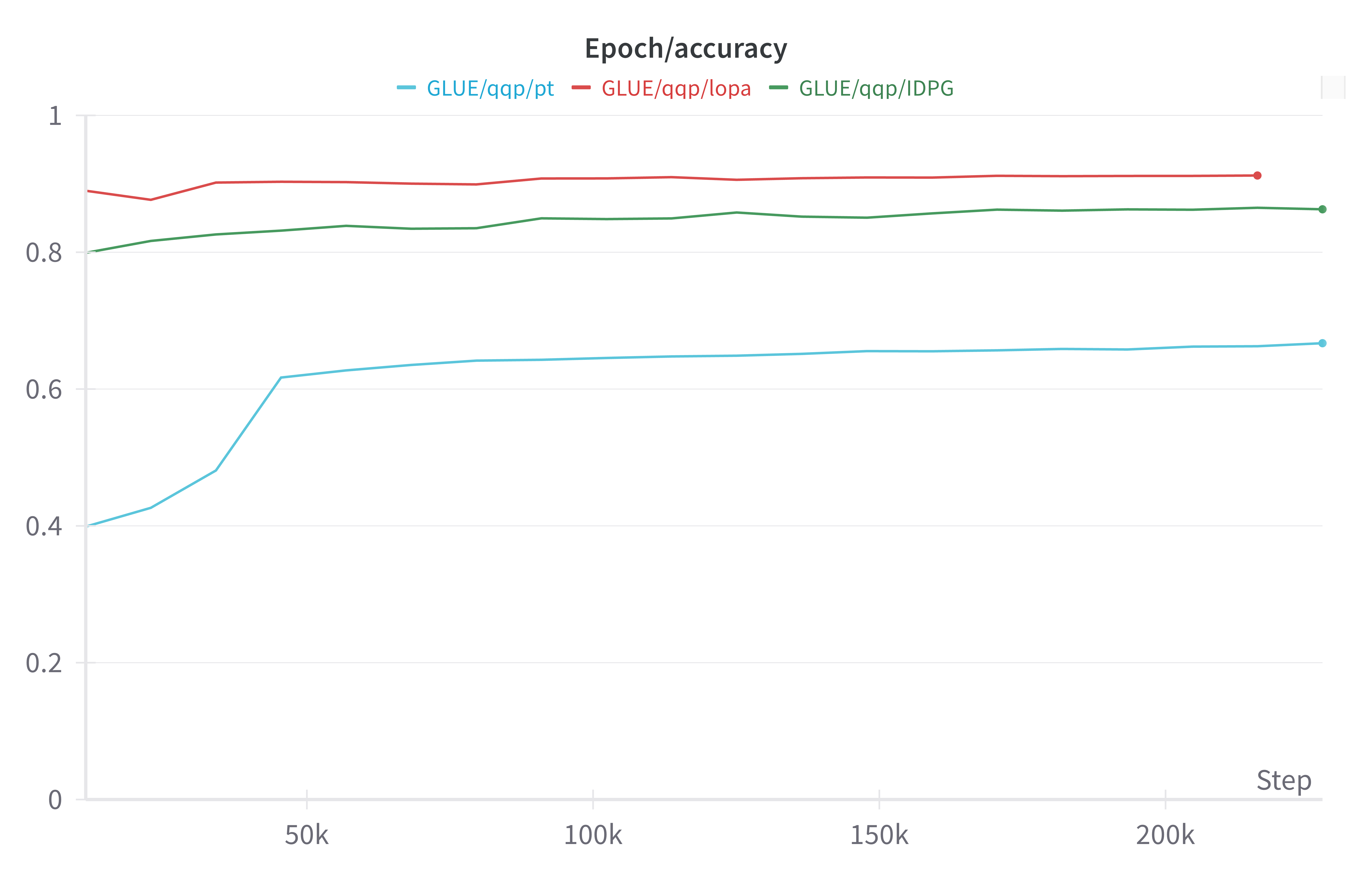}
  }
  \caption{Convergence plots for the PEFT approaches \textcolor{lightblue}{Prompt-tuning (PT)}, \textcolor{darkgreen}{IDPG} and the proposed \textcolor{red}{LOPA} on the NLU task QQP.}
  \label{fig:qqp}
\end{figure}

\begin{figure}
  \centering
  \subfloat[Training Loss]{
    \includegraphics[width=0.48\textwidth]{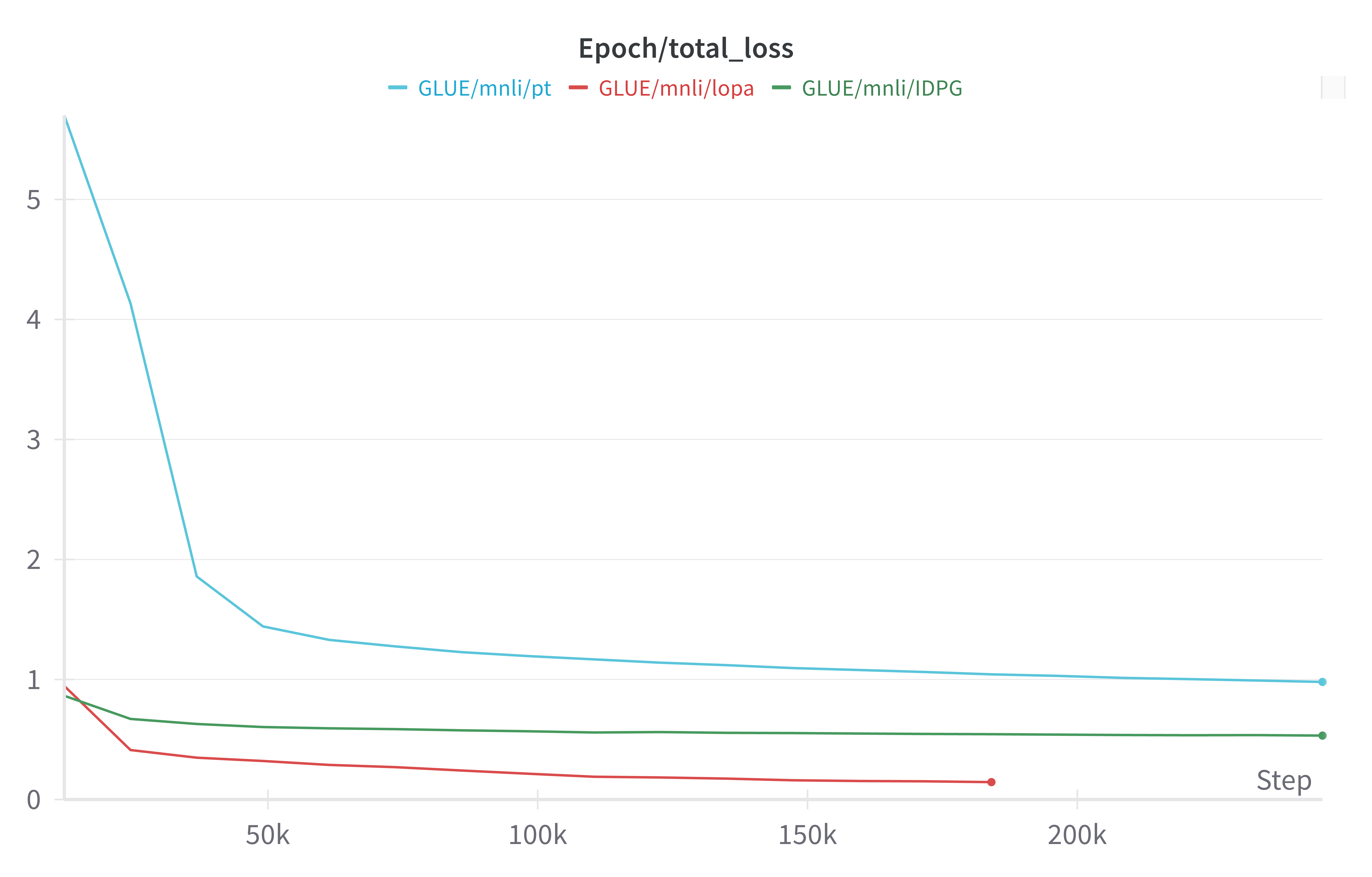}
  }
  \hfill
  \subfloat[Validation Accuracy]{
    \includegraphics[width=0.48\textwidth]{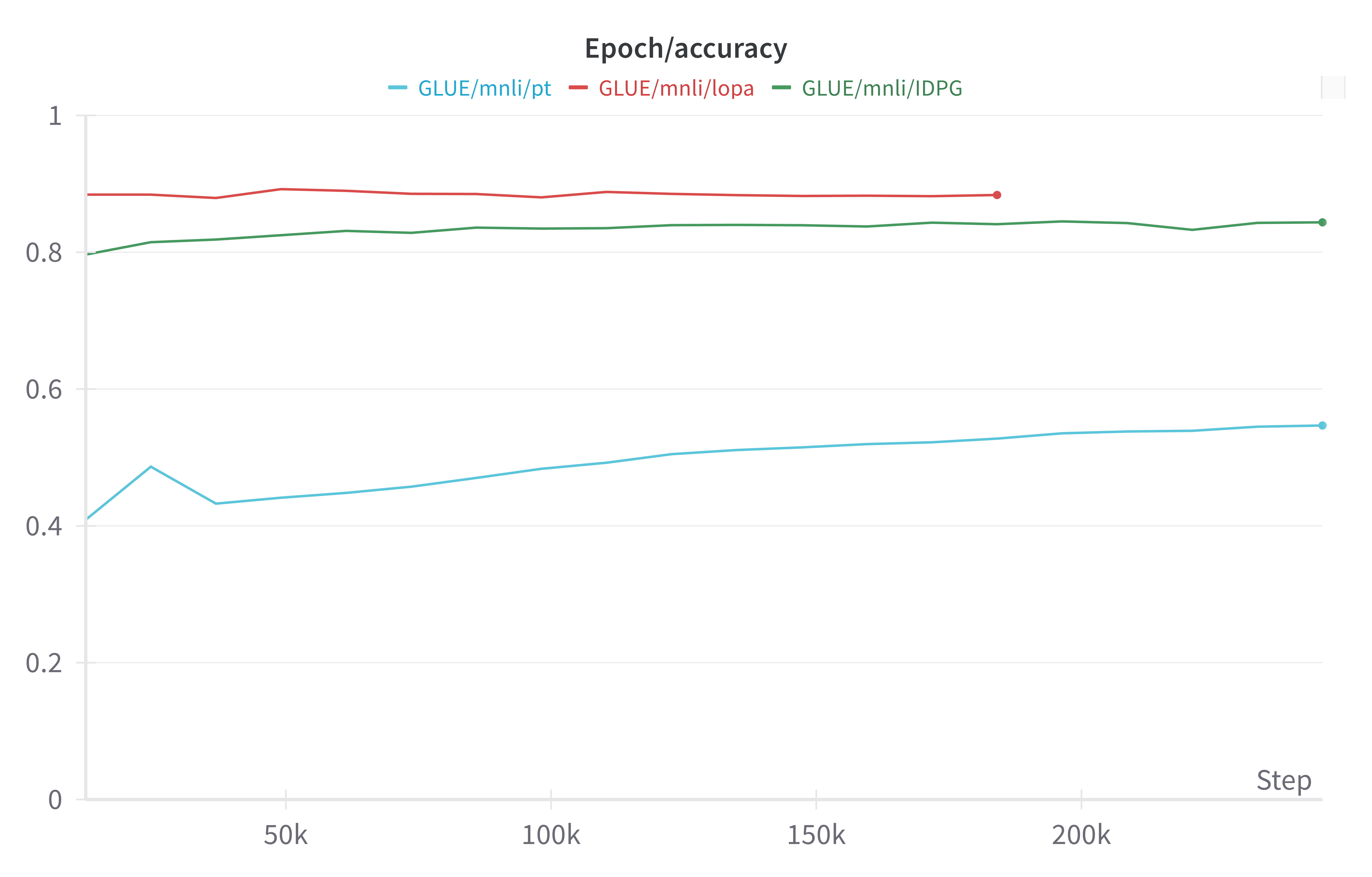}
  }
  \caption{Convergence plots for the PEFT approaches \textcolor{lightblue}{Prompt-tuning (PT)}, \textcolor{darkgreen}{IDPG} and the proposed \textcolor{red}{LOPA} on the NLU task MNLI.}
  \label{fig:mnli}
\end{figure}

\begin{figure}[H]
  \centering
  \subfloat[Training Loss]{
    \includegraphics[width=0.48\textwidth]{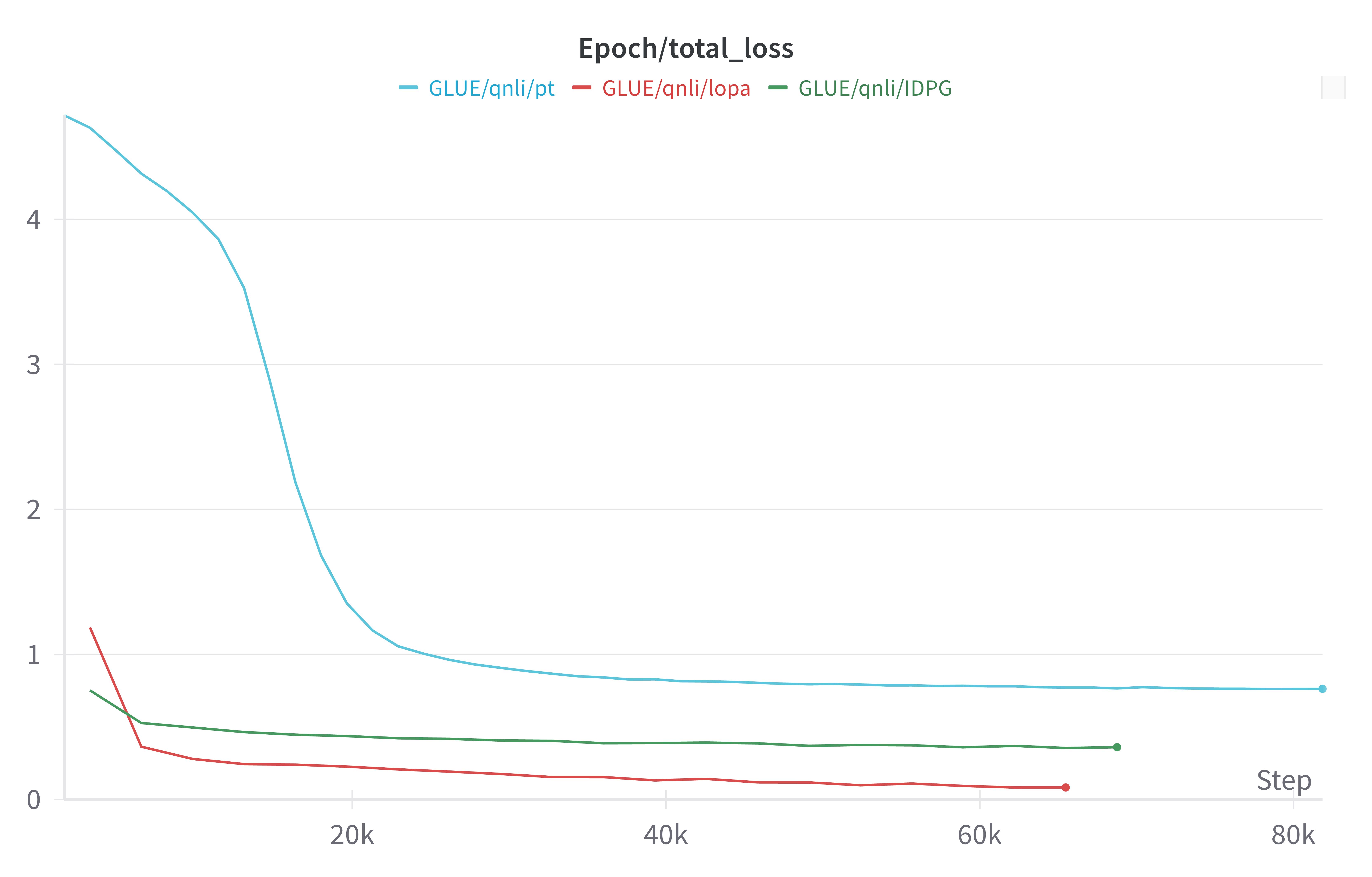}
  }
  \hfill
  \subfloat[Validation Accuracy]{
    \includegraphics[width=0.48\textwidth]{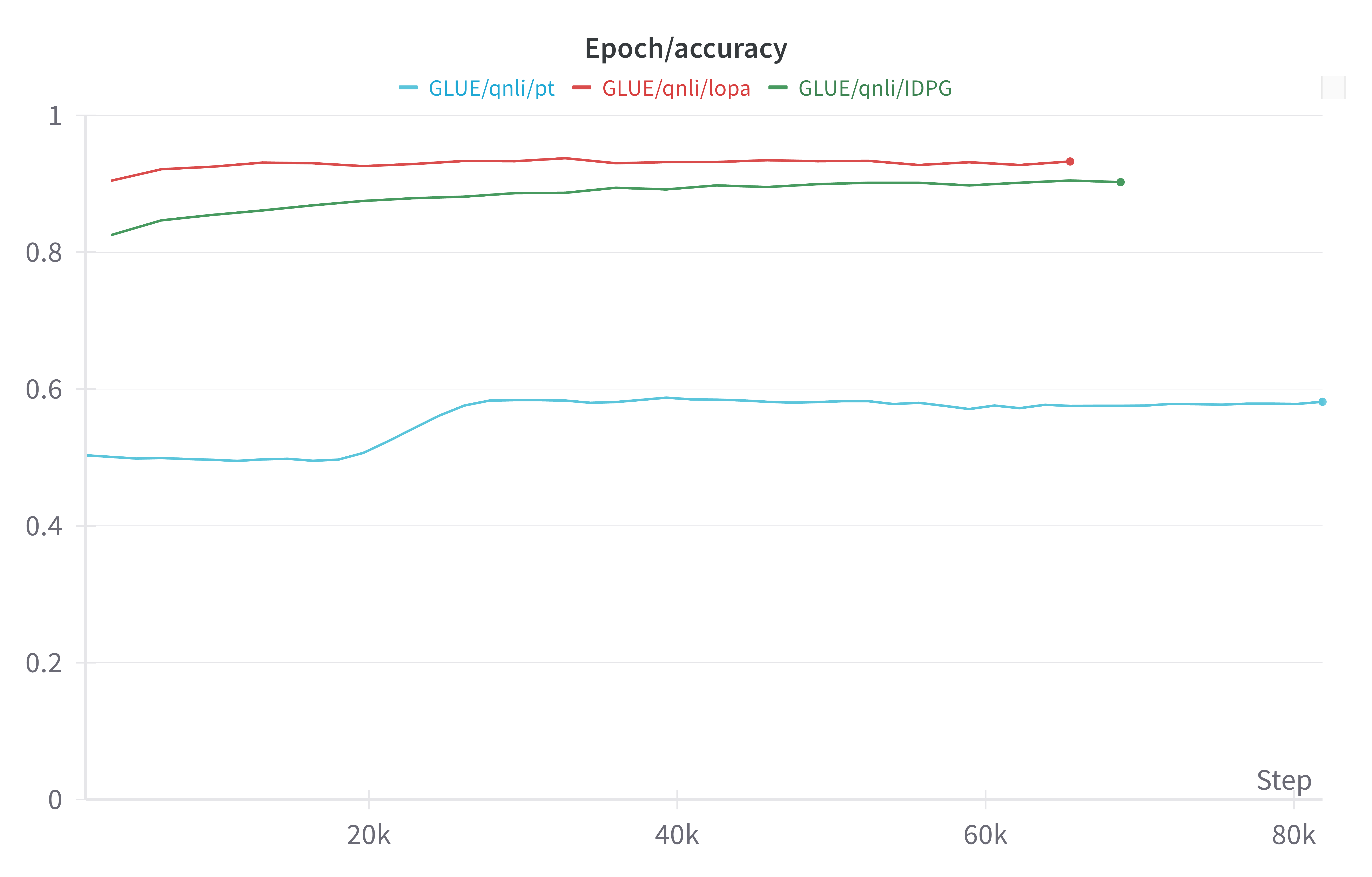}
  }
  \caption{Convergence plots for the PEFT approaches \textcolor{lightblue}{Prompt-tuning (PT)}, \textcolor{darkgreen}{IDPG} and the proposed \textcolor{red}{LOPA} on the NLU task QNLI.}
  \label{fig:qnli}
\end{figure}


\end{document}